\documentclass[letterpaper]{sig-alternate-05-2015}

\usepackage{graphicx} 
\usepackage{subfigure}

\usepackage{algorithm}
\usepackage{algorithmic}

\usepackage{amsmath}
\usepackage{amsfonts}
\usepackage{amssymb}
\usepackage{graphicx}

\renewcommand{\v}[2]{\ensuremath{v_{#2}^{(#1)}}}

\newcommand{\E}{\ensuremath{\mathbb{E}}}

\newcommand{\hh}{\ensuremath{\widehat{h}}}
\newcommand{\bbw}{\ensuremath{\bar{\bw}}}
\newcommand{\bM}{\ensuremath{\bar{M}}}
\newcommand{\bbb}{\ensuremath{\bar{\bb}}}

\newcommand{\field}[1]{\mathbb{#1}}

\newcommand{\bb}{\boldsymbol{b}}

\newcommand{\bx}{\boldsymbol{x}}

\newcommand{\bw}{\boldsymbol{w}}
\newcommand{\bu}{\boldsymbol{u}}

\newcommand{\bv}{\boldsymbol{v}}

\newcommand{\be}{\boldsymbol{e}}
\newcommand{\bzero}{\boldsymbol{0}}

\newcommand{\R}{\field{R}}

\newcommand{\scO}{\mathcal{O}}
\newcommand{\scI}{\mathcal{I}}
\newcommand{\scU}{\mathcal{U}}
\newcommand{\scD}{\mathcal{D}}

\newcommand{\var}{\textsc{var}}

\renewcommand{\tilde}{\widetilde}

\renewcommand{\v}[2]{\ensuremath{\bv_{#2}^{(#1)}}}

\newcommand{\ignore}[1]{}

\DeclareMathOperator*{\argmax}{argmax}

\newcommand{\CB}{\mbox{\sc cb}}

\newtheorem{theorem}{Theorem}

\begin{document}


\CopyrightYear{2016}
\setcopyright{acmcopyright}
\conferenceinfo{SIGIR '16,}{July 17-21, 2016, Pisa, Italy}
\isbn{978-1-4503-4069-4/16/07}\acmPrice{\$15.00}
\doi{http://dx.doi.org/10.1145/2911451.2911548}

\clubpenalty=10000
\widowpenalty = 10000





%

\title{Collaborative Filtering Bandits}
%
%
%
%
%


\numberofauthors{3} 
%
\author{
%
%
\alignauthor
Shuai Li\\
       \affaddr{University of Insubria, Italy}\\
       \email{shuaili.sli@gmail.com}
\alignauthor
Alexandros Karatzoglou\\
       \affaddr{Telef\'onica Research, Spain}\\
       \email{alexk@tid.es}
\alignauthor Claudio Gentile\\
        \affaddr{University of Insubria, Italy}\\
       \email{claudio.gentile@uninsubria.it}
}


\maketitle

\begin{abstract}
Classical collaborative filtering, and content-based filtering methods try to learn a static recommendation model given training data. These approaches are far from ideal in highly dynamic recommendation domains such as news recommendation and computational advertisement, where the set of items and users is very fluid. In this work, we investigate an adaptive clustering technique for content recommendation based on exploration-exploitation strategies in contextual multi-armed bandit settings. Our algorithm takes into account the collaborative effects that arise due to the interaction of the users with the items, by dynamically grouping users based on the items under consideration and, at the same time, grouping items based on the similarity of the clusterings induced over the users. The resulting algorithm thus takes advantage of preference patterns in the data in a way akin to collaborative filtering methods. We provide an empirical analysis on medium-size real-world datasets, showing scalability and increased prediction performance (as measured by click-through rate) over state-of-the-art methods for clustering bandits. We also provide a regret analysis within a standard linear stochastic noise setting.
\end{abstract}


\section{Introduction}\label{s:intro}
Recommender Systems are an essential part of many successful on-line businesses, from e-commerce to on-line streaming, and beyond. Moreover, Computational Advertising can be seen as a recommendation problem where the user preferences highly
depend on the current {\em context}.
In fact, many recommendation domains such as Youtube video recommendation or news recommendation \emph{do not} fit the classical description of a recommendation scenario, whereby a set of users with essentially fixed preferences interact with a fixed set of items. In this classical setting, the well-known {\em cold-start} problem, namely, the lack of accumulated interactions by users on items, needs to be addressed, for instance, by turning to {\em hybrid} recommendation methods~(e.g., \cite{shuai15}). In practice, many relevant recommendation domains are dynamic, in the sense that user preferences and the set of active users change with time. Recommendation domains can be distinguished by how much and how often user preferences and content universe change (e.g., \cite{shuai13}). In highly dynamic recommendation domains, such as news, ads and videos, active users and user preferences are fluid, hence classical collaborative filtering-type methods, such as Matrix or Tensor-Factorization break down. In these settings, it is essential for the recommendation method to adapt to the shifting preference patterns of the users.

Exploration-exploitation methods, a.k.a. multi-armed bandits, have been shown to be an excellent solution for these dynamic domains (see, e.g., the news recommendation evidence in~\cite{lcls10}). While effective, standard contextual bandits do not take collaborative information into account, that is, users who have interacted with similar items in the past will not be deemed to have similar taste based on this fact alone, while items that have been chosen by the same group of users will also not be considered as similar. It is this significant limitation in the current bandit methodology that we try to address in this work.
Past efforts on this problem were based on using online clustering-like algorithms on the graph or network structure of the data in conjunction with multi-armed bandit methods (see Section \ref{s:related}).

Commercial large scale search engines and information retrieval systems are examples of highly dynamic environments where users and items could be described in terms of their membership in some preference cluster. For instance, in a music recommendation scenario, we may have groups of listeners (the users) clustered around music genres, with the clustering changing across different genres. On the other hand, the individual songs (the items) could naturally be grouped by sub-genre or performer based on the fact that they tend to be preferred by the same group of users. Evidence has been collected which suggests that, at least in specific recommendation scenarios, like movie recommendation, data are well modeled by clustering at both user and item sides (e.g., \cite{sst09}).

In this paper, we introduce a Collaborative Filtering based stochastic multi-armed bandit method
that allows for a flexible and generic integration of information of users and items interaction data by alternatively clustering over both user and item sides. Specifically, we describe and analyze an adaptive and efficient clustering of bandit algorithm that can perform collaborative filtering, named {\sc COFIBA} (pronounced as ``coffee bar"). Importantly enough, the clustering performed by our algorithm relies on sparse graph representations, avoiding expensive matrix factorization techniques. We adapt {\sc COFIBA} to the standard setting of sequential content recommendation known as (contextual) multi-armed bandits (e.g.,~\cite{Aue02}) for solving the canonical exploration vs. exploitation dilemma.

Our algorithm works under the assumption that we have to serve content to users in such a way that each content {\em item} determines a clustering over users made up of relatively few groups (compared to the total number of users), within which users tend to react similarly when that item gets recommended. However, the clustering over users need not be the same across different items. Moreover, when the universe of items is large, we also assume that the items might be clustered as a function of the clustering they determine over users, in such a way that the number of {\em distinct} clusterings over users induced by the items is also relatively small compared to the total number of available items.

Our method aims to exploit collaborative effects in a bandit setting in a way akin to the way co-clustering techniques
are used in batch collaborative filtering. Bandit methods also represent one of the most promising approaches to the research community of recommender systems, for instance in tackling the cold-start problem (e.g., \cite{tjll14}), whereby the lack of data on new users leads to suboptimal recommendations. An exploration approach in these cases seems very appropriate.

We demonstrate the efficacy of our dynamic clustering algorithm on three benchmark and real-world datasets.
Our algorithm is scalable and exhibits significant increased prediction performance over the state-of-the-art of clustering bandits. We also provide a regret analysis of the $\sqrt{T}$-style holding with high probability in a standard stochastically linear noise setting.

\section{Learning Model}\label{s:model}
We assume that the user behavior similarity is encoded by a family of
clusterings depending on the specific feature (or context, or item) vector $\bx$ under consideration.
Specifically, we let $\scU = \{1,\ldots, n\}$ represent the set of $n$ users.
Then, given $\bx \in \R^d$, set $\scU$ can be partitioned
into a small number $m(\bx)$ of clusters $U_1(\bx), U_2(\bx), \ldots, U_{m(\bx)}(\bx)$,
where $m(\bx)$ is upper bounded by a constant $m$, independent of $\bx$, with
$m$ being much smaller than $n$. (The assumption $m << n$ is not strictly required but it makes
our algorithms more effective, and this is actually what we expect our datasets to comply with.)
The clusters are such that users belonging to the same cluster $U_j(\bx)$ tend to have similar behavior
w.r.t. feature vector $\bx$ (for instance, they both like or both dislike the item represented by $\bx$), while
users lying in different clusters have significantly
different behavior. The mapping $\bx \rightarrow \{U_1(\bx), U_2(\bx), \ldots, U_{m(\bx)}(\bx)\}$
specifying the actual partitioning of the set of users $\scU$ into the clusters determined by $\bx$
(including the number of clusters $m(\bx)$ and its upper bound $m$), as well as
the common user behavior within each cluster are {\em unknown} to the learning system, and have to be
inferred based on user feedback.

For the sake of simplicity, this paper takes the simple viewpoint that clustering over users is determined
by linear functions $\bx \rightarrow \bu_i^\top\bx$, each one parameterized by an
unknown vector $\bu_i \in \R^d$ hosted at user $i \in \scU$,
in such a way that if users $i$ and $i'$ are in the same cluster w.r.t. $\bx$ then
$\bu_i^\top\bx = \bu_{i'}^\top\bx$, while if $i$ and $i'$ are in different clusters w.r.t.
$\bx$ then $|\bu_i^\top\bx - \bu_{i'}^\top\bx| \geq \gamma$, for some (unknown) gap parameter
$\gamma >0$, independent of $\bx$.\footnote
{
As usual, this assumption may be relaxed by assuming the existence
of two thresholds, one for the within-cluster distance of $\bu_i^\top\bx$ to
$\bu_{i'}^\top\bx$, the other for the between-cluster distance.
}
As in the standard linear bandit setting~(e.g.,
\cite{Aue02,lcls10,chu2011contextual,abbasi2011,cg11,ko11,salso11,yhg12,dkc13,glz14},
and references therein), the unknown vector $\bu_i$ determines the (average) behavior of user $i$.
More concretely, upon receiving context vector $\bx$, user $i$ ``reacts" by delivering a payoff value
\vspace{-0.1in}
\[
a_i(\bx) = \bu_{i}^\top\bx + \epsilon_{i}(\bx)~,
\]
where $\epsilon_{i}(\bx)$ is a conditionally zero-mean and bounded variance noise term so that,
conditioned on the past, the quantity $\bu_{i}^\top\bx$ is indeed the expected payoff observed
at user $i$ for context vector $\bx$.
Notice that the unknown parameter vector $\bu_i$ we associate with user $i$ is supposed to be time invariant
in this model.\footnote
{
It would in fact be possible to lift this whole machinery to time-drifting user preferences by
combining with known techniques (e.g.,~\cite{CCG07,moroshko2015second}).
}

Since we are facing sequential decision settings where the learning system needs to continuously
adapt to the newly received information provided by users, we assume that the learning process
is broken up into a discrete sequence of rounds:
In round $t=1,2,\dots$, the learner receives a user index $i_t \in \scU$ to serve content to,
hence the user to serve may change at every round, though the same user can recur many times.
We assume the sequence of users $i_1, i_2, \ldots$ is determined by an exogenous process that places
nonzero and independent probability to each user being the next one to serve.
Together with $i_t$, the system receives in round $t$ a set of feature vectors
$C_{i_t} = \{\bx_{t,1}, \bx_{t,2},\ldots, \bx_{t,c_t}\} \subseteq \R^d$
encoding the content which is currently available for recommendation to user $i_t$.
The learner is compelled to pick some ${\bar \bx_t} =  \bx_{t,k_t} \in C_{i_t}$
to recommend to $i_t$, and then observes $i_t$'s feedback in the form of
payoff $a_t \in \R$ whose (conditional) expectation is $\bu_{i_t}^\top{\bar \bx_t}$.
The goal of the learning system is to maximize its total payoff\, $\sum_{t=1}^T a_t$\, over $T$ rounds.
When the user feedback at our disposal is only the click/no-click behavior, the payoff $a_t$ is naturally interpreted
as a binary feedback, so that the quantity $\frac{\sum_{t=1}^T a_t}{T}$ becomes a clickthrough rate (CTR), where
$a_t = 1$ if the recommended item was clicked by user $i_t$, and $a_t = 0$, otherwise. CTR is the measure of performance
adopted by our comparative experiments in Section \ref{s:exp}.

From a theoretical standpoint (Section \ref{s:analysis}),
we are instead interested in bounding the cumulative {\em regret} achieved by our algorithms.
More precisely, let the regret $r_t$ of the learner at time $t$ be the extent to which the average
payoff of the best choice in hindsight at user $i_t$ exceeds the average payoff of the algorithm's choice,
i.e.,
\[
r_t = \Bigl(\max_{\bx \in C_{i_t} }\, \bu_{i_t}^\top\bx \Bigl) - \bu_{i_t}^\top{\bar \bx_{t}}~.
\]
We are aimed at bounding with high probability the cumulative regret
\(
\sum_{t=1}^T r_t~,
\)
the probability being over the noise variables $\epsilon_{i_t}({\bar \bx_{t}})$, and any other
possible source of randomness, including $i_t$ -- see Section \ref{s:analysis}.


The kind of regret bound we would like to contrast to is one where
the latent clustering structure over $\scU$ (w.r.t. the feature vectors $\bx$) is somehow known beforehand
(see Section \ref{s:analysis} for details).
When the content universe is large but known a priori, as is frequent in many collaborative
filtering applications, it is often desirable to also group the
items into clusters based on similarity of user preferences, i.e., two items are similar if
they are preferred by many of the same users. This notion of ``two-sided" clustering is well known
in the literature; when the clustering process is simultaneously grouping users based on similarity at the item
side and items based on similarity at the user side, it goes under the name of ``co-clustering''
(see, e.g., \cite{Dhillon:2001,Dhillon:2003}).
Here, we consider a computationally more affordable notion of collaborate filtering based on adaptive two-sided
clustering.
\sloppypar{
Unlike previous existing clustering techniques on bandits~(e.g., \cite{glz14,nl14}),
our clustering setting only applies to the case when the content universe is large but known a priori
(yet, see the end of Section \ref{s:alg}).
Specifically, let the content universe be $\scI = \{\bx_1, \bx_2, \ldots, \bx_{|\scI|}\}$,
and $P(\bx_h) = \{U_1(\bx_h), U_2(\bx_h), \ldots, U_{m(\bx_h)}(\bx_h)\}$ be the partition
into clusters over the set of users $\scU$ induced by item $\bx_h$. Then items $\bx_h, \bx_{h'} \in \scI$ belong
to the same cluster (over the set of items $\scI$) if and only if they induce the same partition of the
users, i.e., if $P(\bx_h) = P(\bx_{h'})$. We denote by $g$ the number of distinct partitions so induced
over $\scU$ by the items in $\scI$, and work under the assumption that $g$ is {\em unknown} but significantly
smaller than $|\scI|$. (Again, the assumption $g << |\scI|$ is not strictly needed, but it both makes
our algorithms effective and is expected to be satisfied in relevant practical scenarios.)
}

Finally, in all of the above, an important special case is when the items to be recommended do not possess
specific features (or do not possess features having significant predictive power).
In this case, it is common to resort to the more classical non-contextual stochastic
multiarmed bandit setting~(e.g., \cite{ACF01,audibert:hal-00711069}), which is recovered
from the contextual framework by setting $d = |\scI|$, and assuming the content universe $\scI$ is
made up of the $d$-dimensional vectors $\be_h, h = 1, \ldots, d$, of the canonical basis of $\R^d$,
As a consequence, the expected payoff
of user $i$ on item $h$ is simply the $h$-th component of vector $\bu_i$, and two users $i$ and $i'$ belong
to the same cluster w.r.t. to $h$ if the $h$-th component of $\bu_i$ equals the $h$-th component of $\bu_{i'}$.
Because the lack of useful annotation on data was an issue with all datasets at our disposal, it is this latter
modeling assumption that motivates the algorithm we actually implemented for the experiments reported in Section \ref{s:exp}.

\section{Related Work}\label{s:related}
Batch collaborative filtering neighborhood methods rely on finding similar groups of users and items to
the target user-item pair, e.g., \cite{vg14}, and thus in effect rely on a dynamic form of grouping users and items.
Collaborative Filtering-based methods have also been integrated with co-clustering techniques,
whereby preferences in each co-cluster are modeled with simple statistics of the preference relations in
the co-cluster, e.g., rating averages \cite{George2005}.

Beyond the general connection to co-clustering~(e.g., \cite{Dhillon:2001,Dhillon:2003}),
our paper is related to the research on multi-armed bandit algorithms for trading off exploration and exploitation through dynamic clustering.
We are not aware of any specific piece of work that combines bandits with co-clustering based on the scheme of collaborative filtering; the papers which are most closely related to ours are \cite{dkc13,mm14,nl14,bcs14,kswa15,glz14,shuai16icml}.
In \cite{dkc13}, the authors work under the assumption that users are defined using a feature vector, and try
to learn a low-rank hidden subspace assuming that variation across users is low-rank. The paper combines low-rank
matrix recovery with high-dimensional Gaussian Process Bandits, but it gives rise to algorithms which do not seem
practical for sizeable problems.
In \cite{mm14}, the authors analyze a non-contextual stochastic bandit problem where model parameters are assumed
to be clustered in a few (unknown) types. Yet, the provided solutions are completely different from ours. The
work \cite{nl14} combines ($k$-means-like) online clustering with a contextual bandit setting, but clustering is only made at
the user side. The paper \cite{bcs14} also relies on bandit clustering at the user side (as in \cite{mm14,nl14}),
with an emphasis on diversifying recommendations to the same user over time.
In \cite{kswa15}, the authors propose cascading bandits of user behavior to identify the $k$ most attractive items,
and formulate it as a stochastic combinatorial partial monitoring problem.
Finally, the algorithms in~\cite{glz14,shuai16icml,gclub} can be seen as a special case of {\sc COFIBA}
when clustering is done only at the user side, under centralized \cite{glz14,gclub} or decentralized \cite{shuai16icml} environments.

Similar in spirit are also \cite{alb13,bl13,cltl15,kbktc15}: In \cite{alb13}, the authors define a transfer learning
problem within a stochastic multi-armed bandit setting, where a prior distribution is defined over
the set of possible models over the tasks; in \cite{bl13}, the authors rely on clustering Markov Decision
Processes based on their model parameter similarity. In \cite{cltl15}, the authors discuss how to choose from $n$
unknown distributions the $k$ ones whose means are largest by a certain metric; in \cite{kbktc15} the authors study
particle Thompson sampling with Rao-Blackwellization for online matrix factorization, exhibiting a regret bound
in a very specific case of $n \times m$ rank-1 matrices. Yet, in none of above cases did the authors make a specific effort towards item-dependent clustering models applied to stochastic multi-armed bandits.

Further work includes \cite{tjll14,tjlzl15}. In \cite{tjll14}, an ensemble of contextual bandits is used to address
the cold-start problem in recommender systems. A similar approach is used in \cite{tjlzl15} to deal with cold-start in recommender systems but based on the probability matching paradigm in a parameter-free bandit strategy, which employs online bootstrap to derive the distribution of the estimated models. In contrast to our work, in neither \cite{tjll14} nor
\cite{tjlzl15} are collaborative effects explicitly taken into account.

\section{The Algorithm}\label{s:alg}
\begin{figure}[t!]
\begin{center}
\begin{picture}(81,170)(81,170)
\scalebox{0.41}{\includegraphics{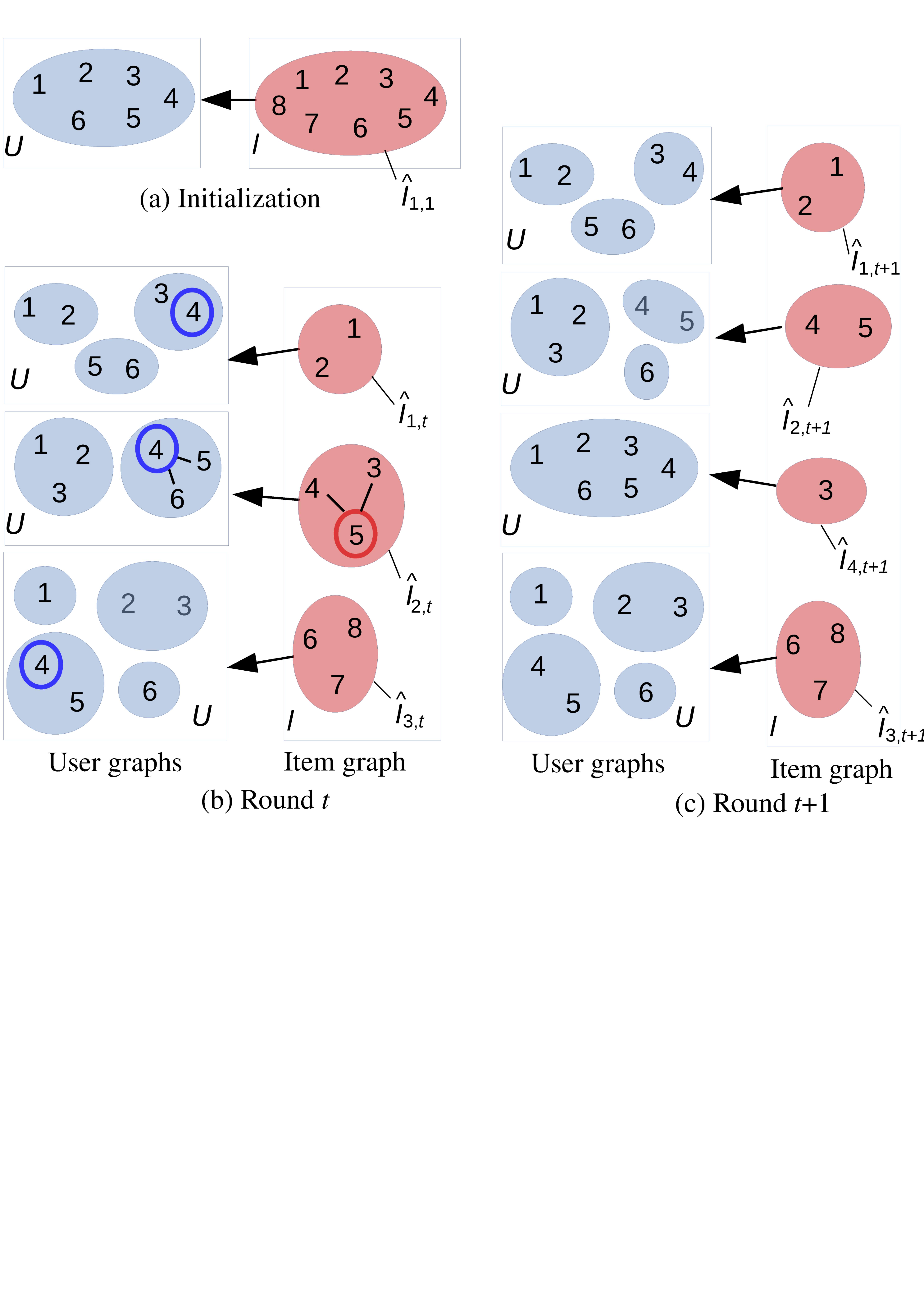}}
\end{picture}
\vspace{0.5in}
\caption{\label{f:cofiba}
In this example, $\scU = \{1,\ldots 6\}$ and $\scI = \{\bx_1,\ldots,\bx_8\}$ (the items are
depicted here as $1, 2, \ldots, 8$).
{\bf (a)} At the beginning we have $g_1 = 1$, with a single item cluster ${\hat I}_{1,1} = \scI$ and, correspondingly,
a single (degenerate) clustering over $\scU$, made up of the unique cluster $\scU$.
{\bf (b)} In round $t$ we have the $g_t = 3$ item clusters $\hat I_{1,t} = \{\bx_1,\bx_2\}$,
$\hat I_{2,t} = \{\bx_3,\bx_4,\bx_5\}$, $\hat I_{3,t} = \{\bx_6,\bx_7,\bx_8\}$.
Corresponding to each one of them are the three clusterings over $\scU$
depicted on the left, so that
$m^U_{t,1} = 3$, $m^U_{t,2} = 2$, and $m^U_{t,3} = 4$.
In this example, $i_t = 4$, and ${\bar \bx_t} = \bx_5$, hence $\hh_t =2$, and we focus on graph $G^U_{t,2}$,
corresponding to user clustering $\{\{1,2,3\},\{4,5,6\}\}$. Suppose in $G^U_{t,2}$ the only neighbors
of user 4 are 5 and 6. When updating such user clustering, the algorithm considers
therein edges $(4,5)$ and $(4,6)$ to be candidates for elimination. Suppose edge $(4,6)$ is eliminated,
so that the new clustering over $\scU$ induced by the updated graph $G^U_{t+1,2}$ becomes
$\{\{1,2,3\},\{4,5\},\{6\}\}$.
After user graph update, the algorithm considers the item graph update.
Suppose $\bx_5$ is only connected to $\bx_4$ and $\bx_3$ in $G^I_t$, and that
$\bx_4$ is not connected to $\bx_3$, as depicted.
Both edge $(\bx_5,\bx_4)$ and edge $(\bx_5,\bx_3)$ are candidates for elimination.
The algorithm computes the neighborhood $N$ of $i_t = 4$ according to $G^U_{t+1,2}$, and compares it
to the the neighborhoods $N^U_{\ell,t+1}(i_t)$, for $\ell = 3,4$. Assume $N \neq N^U_{3,t+1}(i_t)$,
because the two neighborhoods of user 4 are now different, the algorithm deletes edge $(\bx_5,\bx_3)$
from the item graph, splitting the item cluster $\{\bx_3,\bx_4,\bx_5\}$ into the two clusters
$\{\bx_3\}$ and $\{\bx_4,\bx_5\}$, hence allocating a new cluster at the item side corresponding to
a new degenerate clustering $\{\{1,2,3,4,5,6\}\}$ at the user side.
{\bf (c)} The resulting clusterings at the beginning of round $t+1$. (In this picture it is assumed that
edge $(\bx_5,\bx_4)$ was not deleted from the item graph at time $t$.)
}
\end{center}
\vspace{-0.3in}
\end{figure}

{\sc COFIBA}, relies on upper-confidence-based tradeoffs between
exploration and exploitation, combined with adaptive clustering procedures at both the user and the item sides.
{\sc COFIBA} stores in round $t$ an estimate $\bw_{i,t}$ of
vector $\bu_i$ associated with user $i \in \scU$. Vectors $\bw_{i,t}$ are updated based
on the payoff feedback, as in a standard linear least-squares approximation to the corresponding
$\bu_i$. Every user $i \in \scU$ hosts such an algorithm which
operates as a linear bandit algorithm (e.g.,~\cite{chu2011contextual,abbasi2011,glz14})
on the available content $C_{i_t}$.
More specifically, $\bw_{i,t-1}$ is determined by an inverse correlation matrix $M^{-1}_{i,t-1}$
subject to rank-one adjustments, and a vector $\bb_{i,t-1}$ subject to additive updates.
Matrices $M_{i,t}$ are initialized to the $d\times d$ identity matrix,
and vectors $\bb_{i,t}$ are initialized to the $d$-dimensional zero vector.
Matrix $M^{-1}_{i,t-1}$ is also used to define an upper confidence bound $\CB_{i,t-1}(\bx)$
in the approximation of $\bw_{i,t-1}$ to $\bu_i$ along direction $\bx$.
Based on the local information encoded in the weight vectors $\bw_{i,t-1}$ and the confidence bounds $\CB_{i,t-1}(\bx)$,
the algorithm also maintains and updates a family of clusterings of the set of users $\scU$, and a single clustering
over the set of items $\scI$.
On both sides, such clusterings are represented through connected components
of undirected graphs (this is in the same vein as in~\cite{glz14}), where nodes are either users or items.
A pseudocode description of our algorithm is contained in Figures \ref{alg:cofiba}, \ref{alg:userclusterupdate}, and
\ref{alg:itemclusterupdate}, while Figure \ref{f:cofiba} illustrates the algorithm's behavior through a pictorial
example.
\begin{figure}
\begin{center}
\begin{algorithmic}
\vspace{-0.05in}
\small
\STATE \textbf{Input}:
\begin{itemize}
\vspace{-0.07in}
\item Set of users $\scU = \{1,\ldots,n\}$;
\vspace{-0.07in}
\item set of items $\scI =\{\bx_1, \ldots, \bx_{|\scI|}\} \subseteq \R^d$;
\vspace{-0.07in}
\item exploration parameter $\alpha > 0$, and edge deletion parameter $\alpha_2 > 0$.
\end{itemize}
\STATE \textbf{Init}:
\begin{itemize}
\vspace{-0.07in}
\item $\bb_{i,0} = \bzero \in \R^d$ and $M_{i,0} = I \in \R^{d\times d}$,\ \ $i = 1, \ldots n$;
\vspace{-0.07in}
\item {\em User} graph $G^U_{1,1} = (\scU,E^U_{1,1})$, $G^U_{1,1}$ is connected over $\scU$;
\vspace{-0.07in}
\item Number of {\em user} graphs $g_1=1$;
\vspace{-0.05in}
\item No. of {\em user} clusters $m^U_{1,1} = 1$;
\vspace{-0.07in}
\item {\em Item} clusters ${\hat I_{1,1}} = \scI$,\, no. of {\em item} clusters $g_{1} = 1$;
\vspace{-0.07in}
\item {\em Item} graph $G^I_{1} = (\scI,E^I_{1})$, $G^I_{1}$ is connected over $\scI$.
\end{itemize}
\FOR{$t =1,2,\dots,T$}
\STATE Set
\[
\bw_{i,t-1} = M_{i,t-1}^{-1}\bb_{i,t-1}, \qquad i = 1, \ldots, n~;
\]
\STATE Receive $i_t \in \scU$, and get items
       $
       C_{i_t} = \{\bx_{t,1},\ldots,\bx_{t,c_t}  \} \subseteq \scI;
       $
\STATE For each $k=1, \ldots, c_t$, determine which cluster (within the current user clustering
w.r.t. $\bx_{t,k}$) user $i_t$ belongs to, and denote this cluster by $N_k$;
\STATE Compute, for $k=1, \ldots, c_t$, aggregate quantities
\vspace{-0.05in}
\begin{align*}
   \bM_{N_k,t-1}  &= I+\sum_{i \in N_k} (M_{i,t-1}-I),\\[-1mm]
   \bbb_{N_k,t-1} &= \sum_{i \in N_k} \bb_{i,t-1},\\[-1mm]
   \bbw_{N_k,t-1} &= \bM_{N_k,t-1}^{-1}\bbb_{N_k,t-1}~;
\end{align*}
\STATE Set\quad
\vspace{-0.15in}
\[
   k_t = \argmax_{k = 1, \ldots, c_t} \left(\bbw_{N_k,t-1}^\top \bx_{t,k} + \CB_{N_k,t-1}(\bx_{t,k})\right),
\]
where\ \
\(
\CB_{N_k,t-1}(\bx) = \alpha\,\sqrt{\bx^\top\bM_{N_k,t-1}^{-1}\bx\,\log(t+1)}\,;
\)
\STATE Set for brevity ${\bar \bx_t} = \bx_{t,k_t}$;
\STATE Observe payoff $a_t \in \R$, and update weights $M_{i,t}$ and $\bb_{i,t}$ as follows:
       \begin{itemize}
       \vspace{-0.05in}
       \item $M_{i_t,t} = M_{i_t,t-1} + {\bar \bx_t}{\bar \bx_t}^\top$,
       \vspace{-0.05in}
       \item $\bb_{i_t,t} = \bb_{i_t,t-1} + a_t {\bar \bx_t}$,
       \vspace{-0.05in}
       \item Set $M_{i,t} =  M_{i,t-1},\ \bb_{i,t} = \bb_{i,t-1}$ for all $i \neq i_t$~,
       \end{itemize}
\STATE Determine $\hh_t \in \{1,\ldots,g_t\}$ such that $k_t \in {\hat I_{\hh_t,t}}$;
\STATE Update {\em user} clusters at graph $G^U_{t,\hh_{t}} =(\scU,E^U_{t,\hh_{t}})$ by performing the steps in Figure \ref{alg:userclusterupdate};
\STATE For all $h \neq \hh_t$, set $G^U_{t+1,h} = G^U_{t,h}$;
\STATE Update {\em item} clusters at graph $G^I_{t} = (\scI,E^I_{t})$ by performing the steps in Figure \ref{alg:itemclusterupdate}~.
\ENDFOR
\end{algorithmic}
\vspace{-0.1in}
\caption{\label{alg:cofiba}The {\sc COFIBA} algorithm.}
\end{center}
\vspace{-0.25in}
\end{figure}
%

At time $t$, {\sc COFIBA} receives the index $i_t$ of the current user to serve, along with the
available item vectors $\bx_{t,1}, \ldots, \bx_{t,c_t}$, and must select one among them.
In order to do so, the algorithm computes the $c_t$ {\em neighborhood sets} $N_k$, one per item
$\bx_{t,k} \in C_{i_t}$ based on the current aggregation of users (clusters ``at the user side") w.r.t. item $\bx_{t,k}$.
Set $N_k$ should be regarded as the current approximation to the cluster (over the users) $i_t$ belongs to when
the clustering criterion is defined by item $\bx_{t,k}$. Each neighborhood set then defines a compound weight vector
$\bbw_{N_k,t-1}$ (through the aggregation of the corresponding matrices $M_{i,t-1}$ and vectors $\bb_{i,t-1}$) which,
in turn, determines a compound confidence bound\footnote
{
The one given in Figure \ref{alg:cofiba} is the confidence bound we use in our experiments.
In fact, the theoretical counterpart to $\CB$ is significantly more involved,
same efforts can also be found in order to close the gap, e.g., in \cite{audibert:hal-00711069,glz14}.
}
$\CB_{N_k,t-1}(\bx_{t,k})$. Vector $\bbw_{N_k,t-1}$ and confidence bound
$\CB_{N_k,t-1}(\bx_{t,k})$ are combined through an upper-confidence exploration-exploitation scheme so as to
commit to the specific item ${\bar \bx_t} \in C_{i_t}$ for user $i_t$.
Then, the payoff $a_t$ is received, and the algorithm uses ${\bar \bx_t}$ to
update $M_{i_t,t-1}$ to $M_{i_t,t}$ and $\bb_{i_t,t-1}$ to $\bb_{i_t,t}$. Notice that the update is only performed
at user $i_t$, though this will affect the calculation of neighborhood sets and
compound vectors for other users in later rounds.

After receiving payoff $a_t$ and computing $M_{i_t,t}$ and $\bb_{i_t,t}$,
{\sc COFIBA} updates the clusterings at the user side and the (unique) clustering at the item side.
In round $t$, there are multiple graphs $G^U_{t,h} = (\scU,E^U_{t,h})$
at the user side (hence many clusterings over $\scU$, indexed by $h$), and a single graph $G^I_{t} = (\scI,E^I_{t})$
at the item side (hence a single clustering over $\scI$). Each {\em clustering} at the user side
corresponds to a single {\em cluster} at the item side, so that we have $g_t$ {\em clusters}
${\hat I_{1,t}}, \ldots, {\hat I_{g_t,t}}$ over items and $g_t$ {\em clusterings} over users -- see
Figure \ref{f:cofiba} for an example.
\begin{figure}
\begin{center}
\begin{algorithmic}
\small
\STATE Update {\em user} clusters at graph $G^U_{t,\hh_{t}}$ as follows:
\STATE
\begin{itemize}
\item Delete from $E^U_{t,\hh_t}$ all $(i_t,j)$ such that
\vspace{-0.1in}
\[
             |\bw_{i_t,t}^\top{\bar \bx_t} - \bw_{j,t}^\top{\bar \bx_t}|
                 >
               \CB_{i_t,t}({\bar \bx_t}) + \CB_{j,t}({\bar \bx_t})~,
\]
where\ \
      $
        \CB_{i,t}(\bx) = \alpha_2\,\sqrt{\bx^\top M_{i,t}^{-1}\bx\,\log(t+1)}\,;
      $
\vspace{-0.1in}
\item Let $E^U_{t+1,\hh_t}$ be the resulting set of edges, set $G^U_{t+1,\hh_t} = (\scU,E^U_{t+1,\hh_t})$,
and compute associated clusters $\hat U_{1,t+1,\hh_t}, \hat U_{2,t+1,\hh_t}, \ldots, \hat U_{m^U_{t+1,\hh_t},t+1,\hh_t}$
as the connected components of $G^U_{t+1,\hh_t}$.
\end{itemize}
\end{algorithmic}
\vspace{-0.05in}
\caption{\label{alg:userclusterupdate}User cluster update in the {\sc COFIBA}}
\end{center}
\vspace{-0.1in}
\end{figure}
On both user and item sides, updates take the form of edge deletions. Updates at the user side are only performed
on the graph $G^U_{t,\hh_t}$ pointed to by the selected item ${\bar \bx_t} = \bx_{t,k_t}$.
Updates at the item side are only made if it is likely that the neighborhoods of user $i_t$
has significantly changed when considered w.r.t. two previously deemed similar items.
Specifically, if item $\bx_h$ was directly connected to item ${\bar \bx_t}$ at the beginning of
round $t$ and, as a consequence of edge deletion at the user side, the set of users that are now
likely to be close to $i_t$ w.r.t. $\bx_h$ is no longer the same as the set of users that are likely
to be close to $i_t$ w.r.t. ${\bar \bx_t}$, then this is taken as a good indication that
item $\bx_h$ is not inducing the same partition over users as ${\bar \bx_t}$ does, hence
edge $({\bar \bx_t}, \bx_h)$ gets deleted.
Notice that this need not imply that, as a result of this deletion, the two items are now
belonging to different clusters over $\scI$, since these two items may still be indirectly connected.

\begin{figure}
\begin{center}
\begin{algorithmic}
\small
\STATE Update {\em item} clusters at graph $G^I_{t}$ as follows:
\STATE
\begin{itemize}
\item
For all $\ell$ such that $({\bar \bx_t},\bx_{\ell}) \in E^I_{t}$
build neighborhood $N^U_{\ell,t+1}(i_t)$ as:
\vspace{-0.1in}
\begin{align*}
N^U_{\ell,t+1}(i_t) = \Bigl\{j\,:\,j& \neq i_t\,, | \bw_{i_t,t}^\top\bx_{\ell} - \bw_{j,t}^\top\bx_{\ell}| \\
                & \leq
               \CB_{i_t,t}(\bx_{\ell}) + \CB_{j,t}(\bx_{\ell}) \Bigl\}\,;
\end{align*}
\vspace{-0.2in}
\item Delete from $E^I_{t}$ all $({\bar \bx_t},\bx_{\ell})$ such that $N^U_{\ell,t+1}(i_t) \neq N^U_{k_t,t+1}(i_t)$,
where $N^U_{k_t,t+1}(i_t)$ is the {\em neighborhood} of node $i_t$ w.r.t. graph $G^U_{t+1,\hh_t}$\,;
\vspace{-0.1in}
\item Let $E^I_{t+1}$ be the resulting set of edges, set $G^I_{t+1} = (\scI,E^I_{t+1})$, compute associated
{\em item} clusters $\hat I_{1,t+1}, \hat I_{2,t+1}, \ldots, \hat I_{g_{t+1},t+1}$~ through the connected components
of $G^I_{t+1}$;
\vspace{-0.05in}
\item For each new item cluster created,
allocate a new connected graph over users representing a single (degenerate) cluster $\scU$.
\end{itemize}
\end{algorithmic}
\caption{\label{alg:itemclusterupdate}Item cluster update in the {\sc COFIBA}}
\end{center}
\vspace{-0.25in}
\end{figure}
\vspace{0.5in}

\vspace{-0.5in}
It is worth stressing that a naive implementation of {\sc COFIBA} would require memory allocation for maintaining
$|\scI|$-many $n$-node graphs, i.e., $\scO(n^2\,|\scI|)$. Because this would be prohibitive even
for moderately large sets of users, we make full usage of the approach of~\cite{glz14}, where
instead of starting off with complete graphs over users each time a new cluster over items is created,
we randomly sparsify the complete graph by drawing an Erdos-Renyi initial graph, still retaining
with high probability the underlying clusterings
$\{U_1(\bx_h),\ldots, U_{m(\bx_h)}(\bx_h)\}$, $h = 1,\ldots, |\scI|$, over users.
This works under the assumption that
the latent clusters $U_i(\bx_h)$ are not too small -- see the argument
in~\cite{glz14}, where it is shown that in practice the initial graphs can have $\scO(n\log n)$ edges
instead of $\scO(n^2)$. Moreover, because we modify the item graph by edge deletions only, one can
show that with high probability (under the modeling assumptions of Section \ref{s:model})
the number $g_t$ of clusters over items remains upper bounded by $g$
throughout the run of {\sc COFIBA}, so that the actual storage required by the algorithm
is indeed $\scO(ng\log n)$. This also brings a substantial saving in running time, since updating
connected components scales with the number of edges of the involved graphs. It is this graph sparsification
techniques that we used and tested along the way in our experimentation parts.

Finally, despite we have described in Section \ref{s:model} a setting where $\scI$ and $\scU$ are
known a priori (the analysis in Section \ref{s:analysis} currently holds only in this scenario),
nothing prevents in practice to adapt {\sc COFIBA} to the case when new content or new users show up.
This essentially amounts to adding new nodes to the graphs at either the item or the user side, by
maintaining data-structures via dynamic memory allocation. In fact, this is precisely how we implemented our
algorithm in the case of very big item or user sets (e.g., the Telefonica and the Avazu dataset in the next section).

\section{Experiments}\label{s:exp}
We compared our algorithm to standard bandit baselines on three real-world datasets: one canonical benchmark dataset on news recommendations, one advertising dataset from a living production system, and one publicly available advertising dataset. In all cases, no features on the items have been used. We closely followed the same experimental setting as in previous work
~\cite{chu2011contextual,glz14}, thereby evaluating prediction performance by click-through rate.

\subsection{Datasets}
\textbf{Yahoo!.} The first dataset we use for the evaluation is the freely available benchmark dataset which
was released in the ``ICML 2012 Exploration \& Exploitation Challenge"\footnote{https://explochallenge.inria.fr/category/challenge}. The aim of the challenge was to build state-of-the-art news article recommendation algorithms on Yahoo! data, by building an algorithm that learns efficiently a policy to serve news articles on a web site.
The dataset is made up of random traffic records of user visits on the ``Today Module'' of Yahoo!, implying that both the visitors and the recommended news article are selected randomly.
The available options (the items) correspond to a set of news articles available for recommendation, one being displayed
in a small box on the visited web page. The aim is to recommend an interesting article to
the user, whose interest in a given piece of news is asserted by a click on it.
The data has $30$ million visits over a two-week time stretch. Out of the logged information contained in each record,
we used the user ID in the form of a $136$-dimensional boolean vector containing his/her features (index $i_t$), the
set of relevant news articles that the system can recommend from (set $C_{i_t}$); a randomly recommended article during the visit; a boolean value indicating whether the recommended article was clicked by the visiting user or not (payoff $a_t$).
Because the displayed article is chosen uniformly at random from the candidate article pool, one can use an unbiased off-line evaluation method to compare bandit algorithms in a reliable way.
We refer the reader to ~\cite{glz14} for a more detailed description of how this dataset was
collected and extracted. We picked the larger of the two datasets considered in \cite{glz14}, resulting in
$n \approx 18K$ users, and $d = 323$ distinct items.
The number of records ended up being $2.8M$, out of which we took the first $300K$ for parameter tuning,
and the rest for testing.


\textbf{Telefonica.} This dataset was obtained from Telefonica S.A., which is the number one Spanish broadband and
telecommunications provider, with business units in Europe and South America. This data contains clicks on ads displayed to user on one of the websites that Telefonica operates on. The data were collected from the back-end server logs, and
consist of two files: the first file contains the ads interactions (each record containing an impression timestamp, a user-ID, an action, the ad type, the order item ID, and the click timestamp); the second file contains the ads metadata as item-ID, type-ID, type, order-ID, creative type, mask, cost, creator-ID, transaction key, cap type.
Overall, the number $n$ of users was in the scale of millions, while the number $d$ of items was approximately 300. The data contains $15M$ records, out of which we took the first $1,5M$ for parameter tuning, and the rest for testing.
Again, the only available payoffs are those associated with the items served by the system. Hence, in order to make the
procedure be an effective estimator in a sequential decision process (e.g.,~\cite{cg11,dudik12,glz14,lcls10}), we {\em simulated} random choices by the system by generating the available item sets $C_{i_t}$ as follows:
At each round $t$, we stored the ad served to the current user $i_t$ and the associated payoff value $a_t$
(1 =``clicked", 0 =``not clicked"). Then we created
$C_{i_t}$ by including the served ad along with $9$ extra items (hence $c_t = 10$ $\forall t$) which
were drawn uniformly at random in such a way that, for any item $\be_h \in \scI$, if $\be_h$ occurs in some set
$C_{i_t}$, this item will be the one served by the system $1/10$ of the times. The
random selection was done independent of the available payoff values $a_t$.
All our experiments on this dataset were run on a machine with $64$GB RAM and $32$ Intel Xeon cores.

\textbf{Avazu.} This dataset was prepared by Avazu Inc,\footnote{https://www.kaggle.com/c/avazu-ctr-prediction} which is a leading multinational corporation in the digital advertising business. The data was provided for the challenge to predict the click-through rate of impressions on mobile devices, i.e., whether a mobile ad will be clicked or not. The number of samples was around $40M$, out of which we took the first $4M$ for parameter tuning, and the remaining for testing. Each line in the data file represents the event of an ad impression on the site or in a mobile application (app), along with additional context information. Again, payoff $a_t$ is binary.
The variables contained in the dataset for each sample are the following: ad-ID; timestamp (date and hour); click (boolean variable); device-ID; device IP; connection type; device type; ID of visited App/Website; category of visited App/Website; connection domain of visited App/Website; banner position; anonymized categorical fields (C1, C14-C21). We pre-processed the dataset as follows: we cleaned up the data by filtering out the records having missing feature values,
and removed outliers.
We identified the user with device-ID, if it is not null. The number of users on this dataset is in the scale of millions. Similar to the Telefonica dataset, we generated recommendation lists of length $c_t = 20$ for each distinct timestamp. We used the first $4M$ records for tuning parameters, and the remaining $36M$ for testing.
All data were transferred to Amazon S3, and all jobs were run through the Amazon EC2 Web Service.

\subsection{Algorithms}
We compared {\sc COFIBA} to a number of state-of-the-art bandit algorithms:
\begin{itemize}
\vspace{-0.03in}
\item {\sc LINUCB-ONE} is a single instance of the {\sc ucb1}~\cite{ACF01} algorithm, which is a very popular and established algorithm that has received a lot of attention in the research community over the past years;
\vspace{-0.03in}
\item {\sc DYNUCB} is the dynamic {\sc ucb} algorithm of~\cite{nl14}. This algorithm adopts a ``$K$-means''-like
clustering technique so as to dynamically re-assign the clusters on the fly based on the changing
contexts and user preferences over time;
\vspace{-0.03in}
\item {\sc LINUCB-IND}~\cite{glz14} is a set of independent {\sc ucb1} instances, one per user, which provides a fully personalized recommendation for each user;
\vspace{-0.03in}
\item {\sc CLUB}~\cite{glz14} is the state-of-the-art online clustering of bandits algorithm that dynamically cluster users based on the confidence
ellipsoids of their models;
\vspace{-0.03in}
\item {\sc LINUCB-V}~\cite{audibert:hal-00711069} is also a single instance of {\sc ucb1}, but with a more sophisticated confidence bound; this algorithm turned out to be
the winner of the ``ICML 2012 Challenge'' where the Yahoo! dataset originates from.
\end{itemize}
We tuned the optimal parameters in the training set with a standard grid search as indicated in~\cite{cg11,glz14}, and used the test set to evaluate the predictive performance of the algorithms. Since the system's recommendation need not coincide with the recommendation issued by the algorithms we tested, we only retained the records on which the two recommendations were indeed the same. Because records are discarded on the fly, the actual number $T$ of retained records (``Rounds" in the plots of the next subsection) changes slightly across algorithms; $T$ was around $70K$ for the Yahoo! data, $350K$ for the Telefonica data, and $900K$ for the Avazu data. All experimental results we report were averaged over $3$ runs (but in fact the variance
we observed across these runs was fairly small).

\subsection{Results}
\begin{figure}[t*]
\begin{picture}(0,105)(0,105)
\includegraphics[scale=0.4]{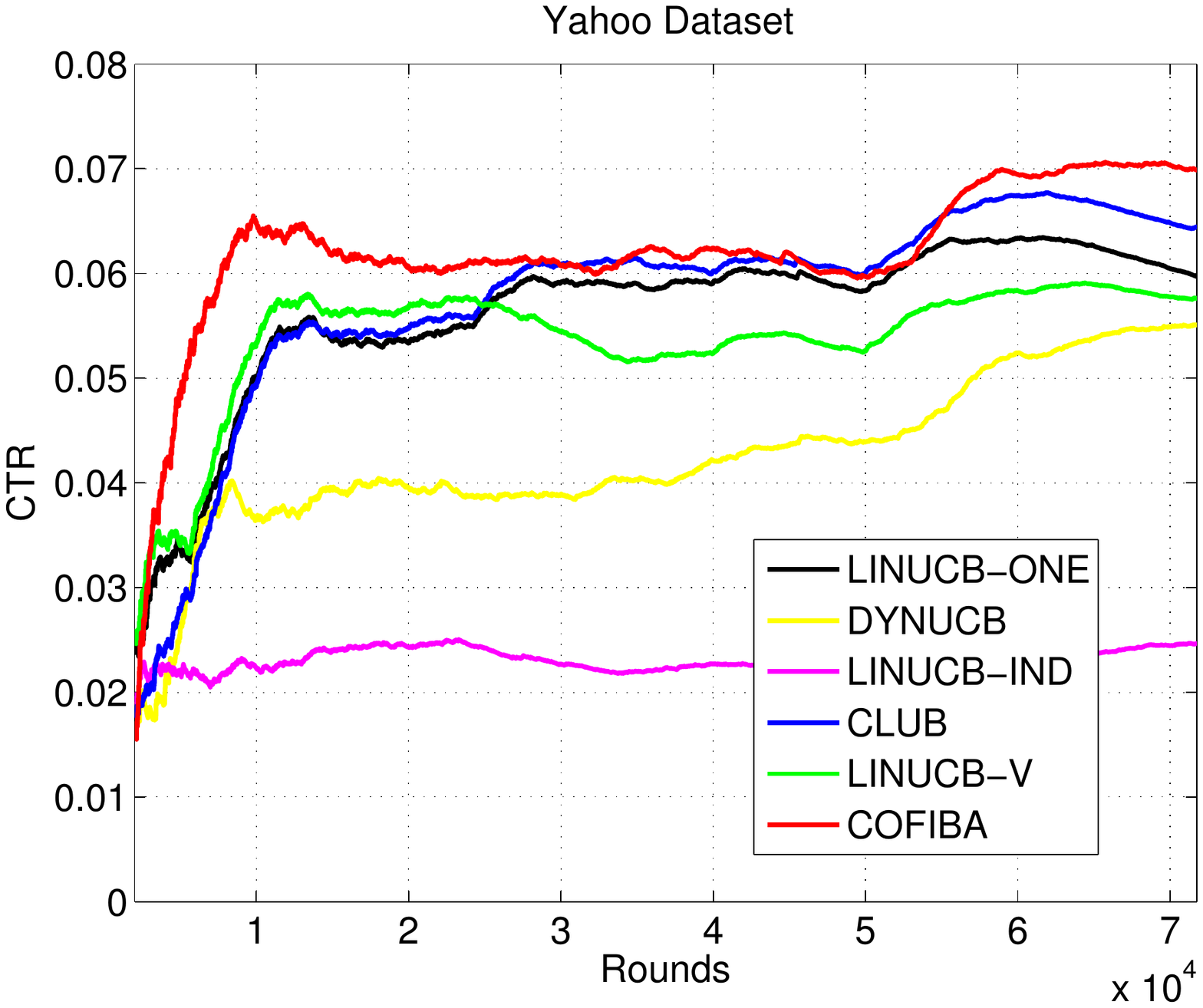}
\end{picture}
\vspace{0.45in}
\caption{\label{fig:yahoo}Results on the Yahoo dataset.
}
\vspace{-0.1in}
\end{figure}

\begin{figure}[t*]
\begin{picture}(0,105)(0,105)
\includegraphics[scale=0.41]{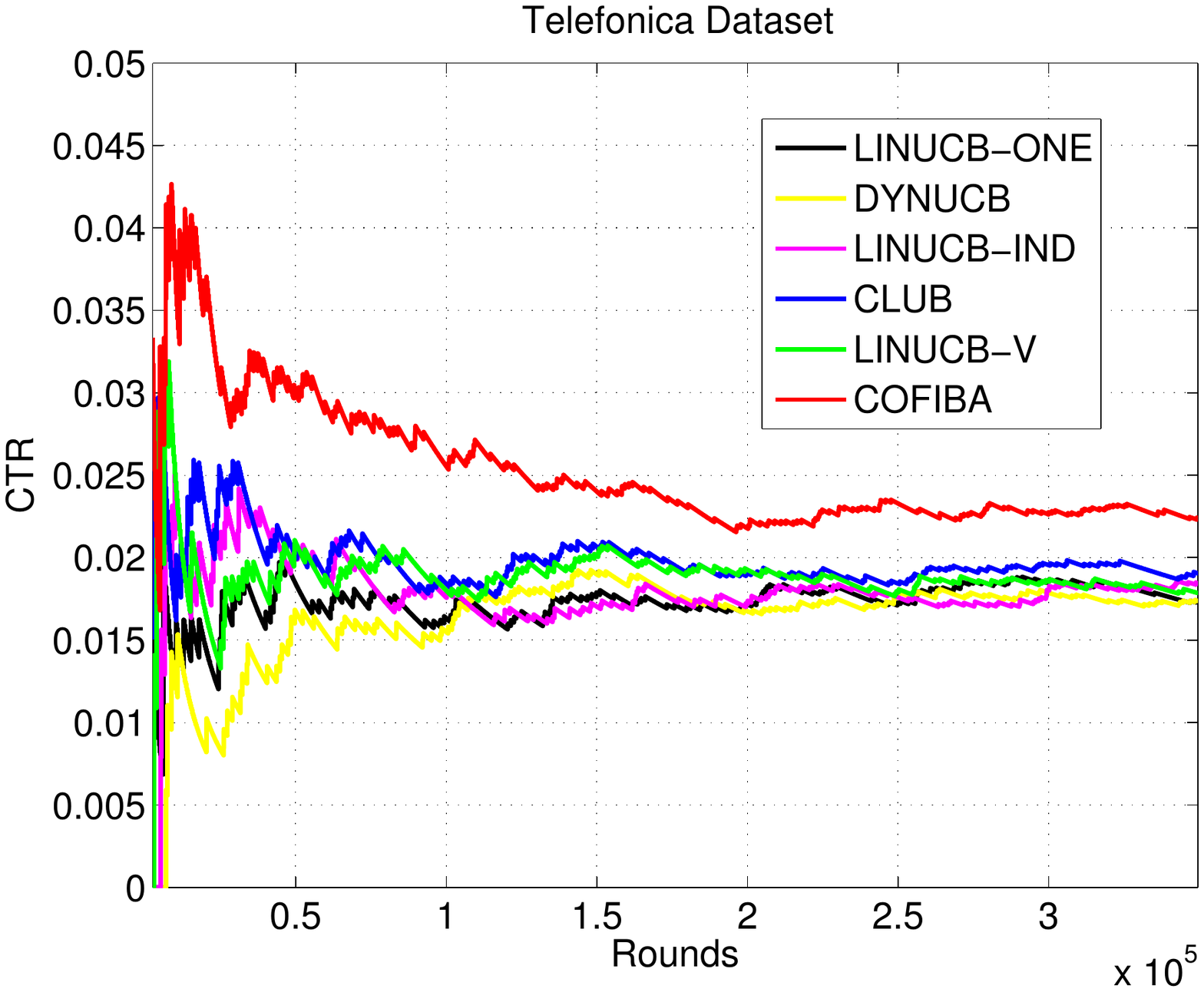}
\end{picture}
\vspace{0.45in}
\caption{\label{fig:telefonica}Results on the Telefonica dataset.
}
\vspace{0.5in}
\end{figure}

\begin{figure}[t*]
\begin{picture}(0,110)(0,110)
\includegraphics[scale=0.41]{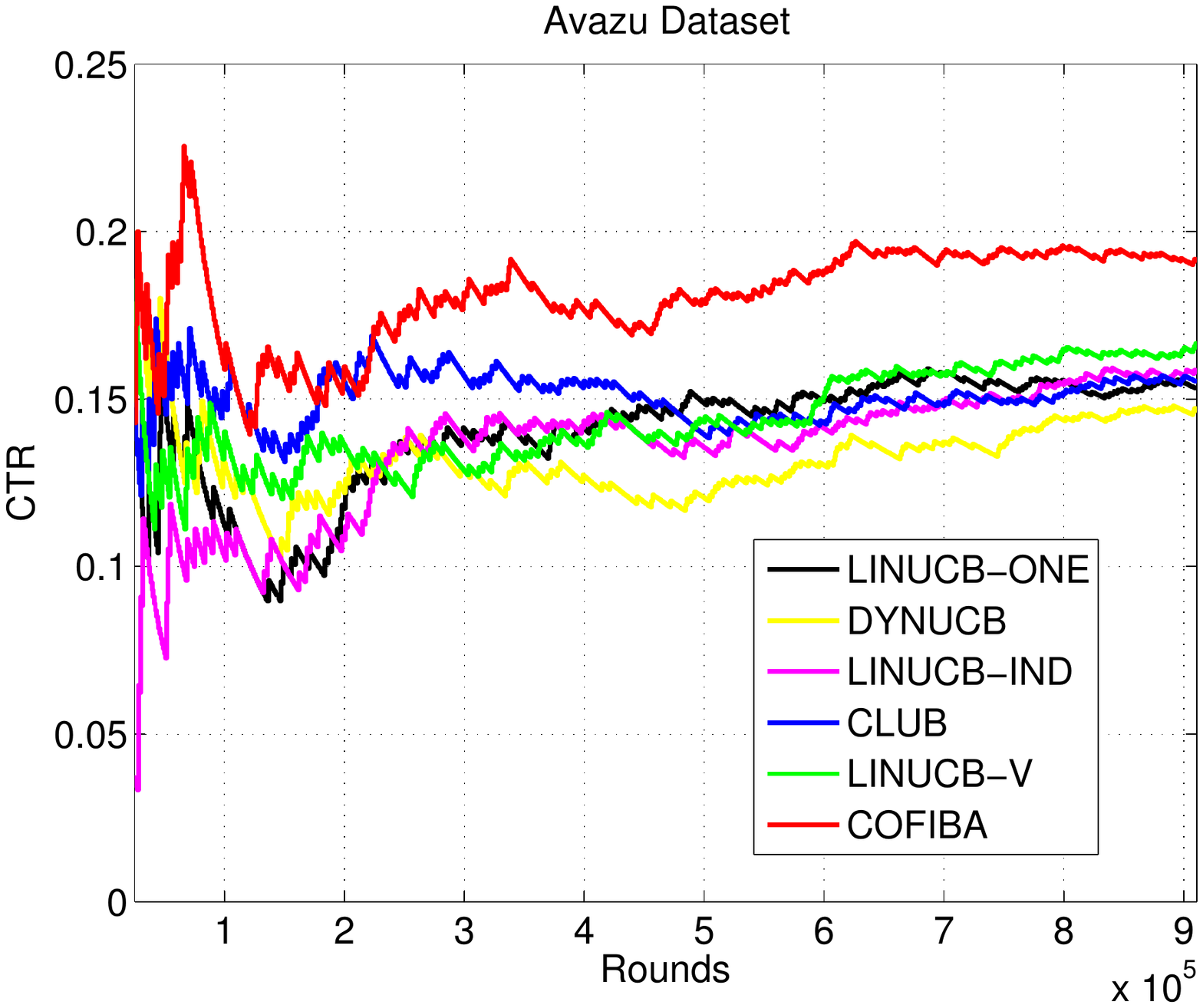}
\end{picture}
\vspace{0.45in}
\caption{\label{fig:avazu}Results on the Avazu dataset.
}
\vspace{-0.1in}
\end{figure}

\begin{figure}[t*]
\begin{picture}(0,52)(0,52)
\begin{tabular}{l}
\includegraphics[width=0.45\textwidth]{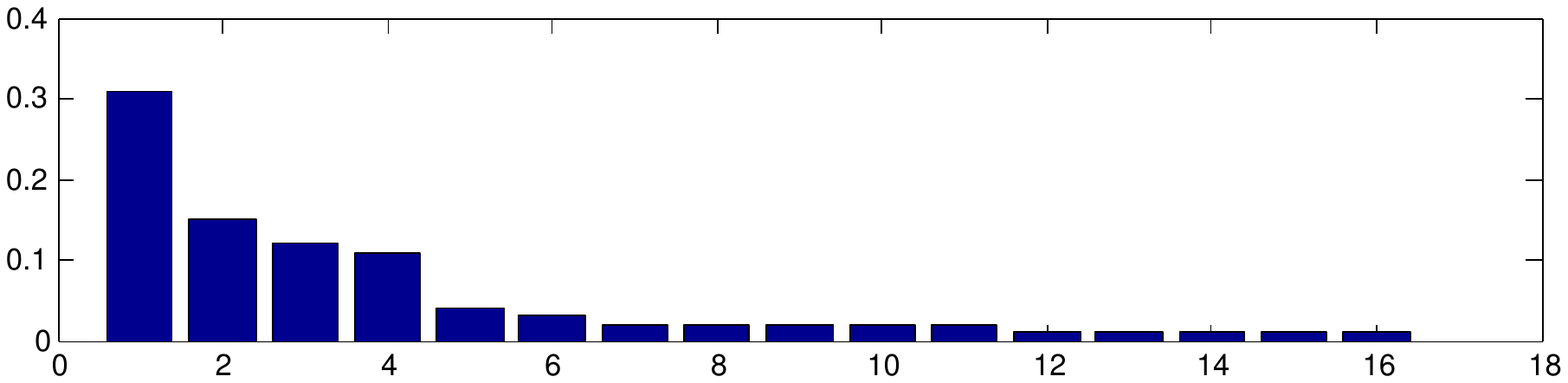}\\[-3.44in]
\includegraphics[width=0.45\textwidth]{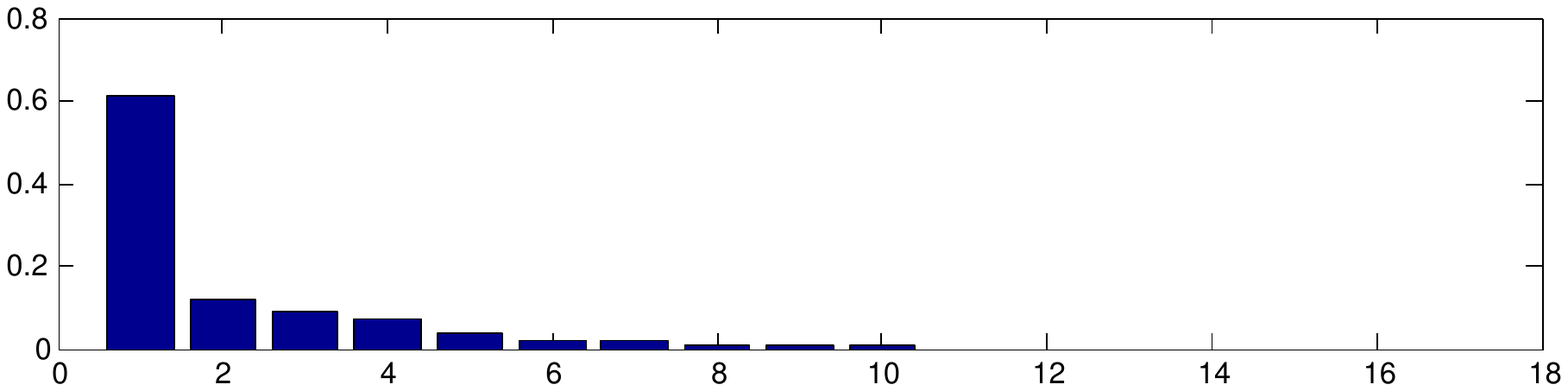}\\[-3.44in]
\includegraphics[width=0.45\textwidth]{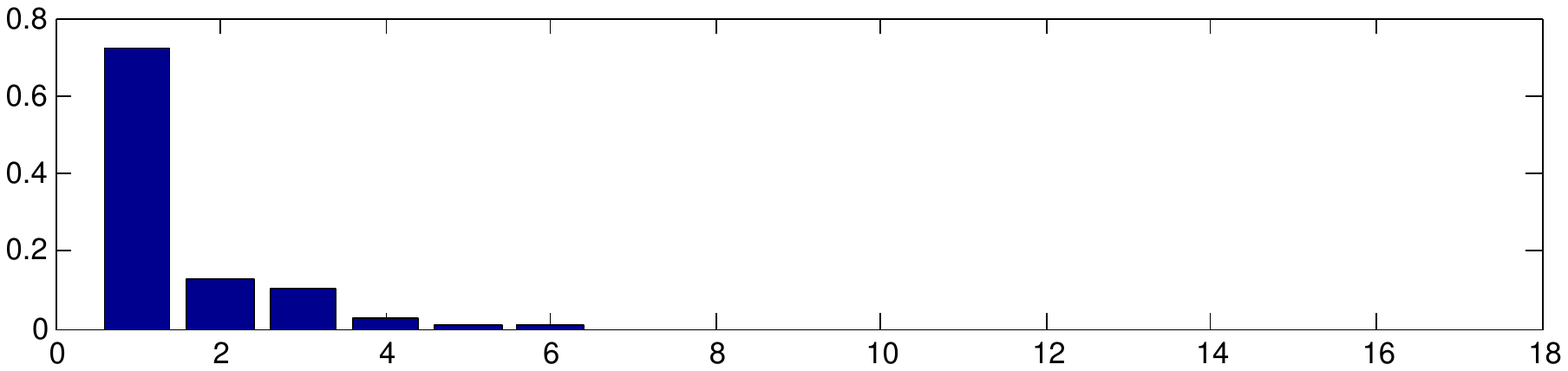}\\[-3.40in]
\includegraphics[width=0.45\textwidth]{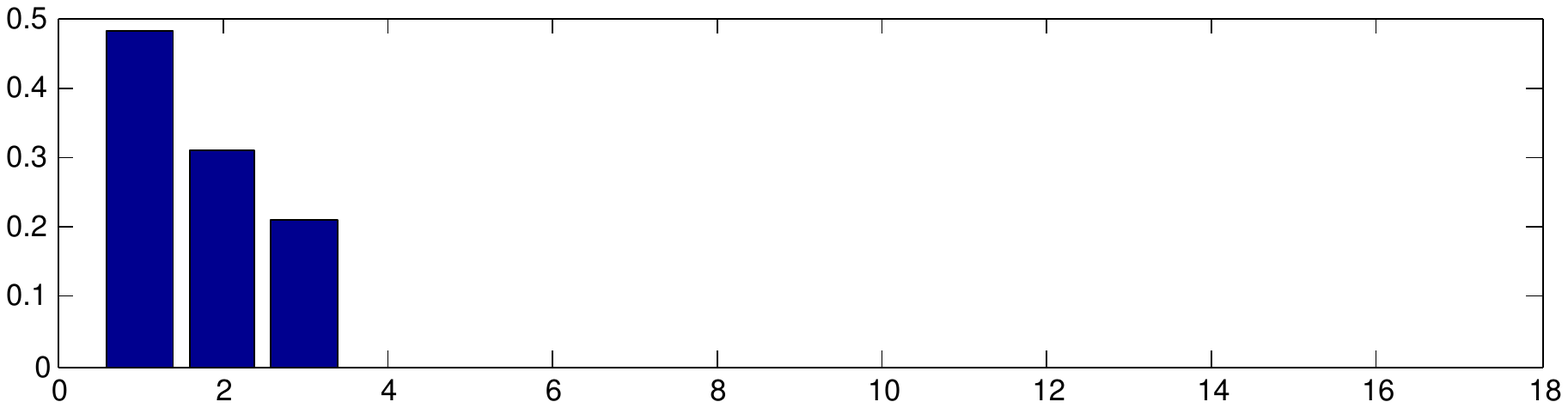}\\[-3.34in]
\includegraphics[width=0.45\textwidth]{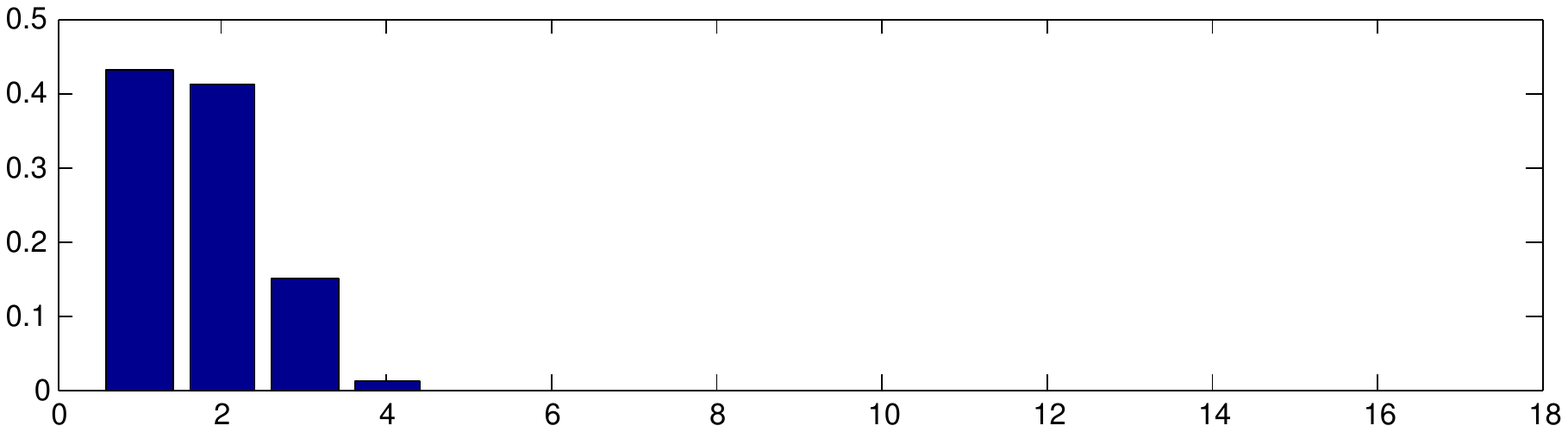}
\end{tabular}
\end{picture}
\vspace{2.8in}
\caption{\label{fig:bars}A typical distribution of cluster sizes over users for the Yahoo dataset. Each bar plot corresponds
to a cluster at the item side. We have 5 plots since this is the number of clusters over the items that {\sc COFIBA} ended up
with after sweeping once over this dataset in the run at hand.
Each bar represents the fraction of users contained in the corresponding cluster.
For instance, the first cluster over the items generated 16 clusters over the users (bar plot on top), with relative sizes 31\%,
15\%, 12\%, etc. The second cluster over the items generated 10 clusters over the users (second bar plot from top) with relative
sizes 61\%, 12\%, 9\%, etc. The relative size of the 5 clusters over the items is as follows: 83\%, 10\%, 4\%, 2\%, and 1\%, so that the clustering pattern depicted in the top plot applies to 83\% of the items, the second one to 10\% of the items, and so on.
}
\vspace{-0.1in}
\end{figure}

Our results are summarized in Figures \ref{fig:yahoo}, \ref{fig:telefonica}, and \ref{fig:avazu}.
Further evidence is contained in Figure \ref{fig:bars}.
In Figures \ref{fig:yahoo}--\ref{fig:avazu}, we plotted click-through rate (``CTR") vs. retained records so
far (``Rounds"). All these experiments are aimed at testing the performance of the various bandit algorithms
in terms of prediction performance, also in cold-start regimes (i.e., the first relatively small
fraction of the time horizon in the $x$-axis). Our experimental setting is in line with previous
ones (e.g., \cite{chu2011contextual,glz14}) and, by the way the data have been prepared, gives rise to a reliable
estimation of actual CTR behavior under the same experimental conditions as in~\cite{chu2011contextual,glz14}.
Figure \ref{fig:bars} is aimed at supporting the theoretical model of Section
\ref{s:model}, by providing some evidence on the kind of clustering statistics produced by {\sc COFIBA} at the end
of its run.

Whereas the three datasets we took into consideration are all generated by real online web applications, it is worth pointing
out that these datasets are indeed different in the way customers consume the associated content.
Generally speaking, the longer the lifecycle of one item the fewer the items, the higher the chance that users with similar preferences will consume it, and hence the bigger the collaborative effects contained in the data. It is therefore reasonable to expect that our algorithm will be more effective in datasets where the collaborative effects are indeed strong.

The users in the Yahoo! data (Figure \ref{fig:yahoo}), are likely to span a wide range of demographic characteristics; on top of this, this dataset is derived from the consumption of news that are often interesting for
large portions of these users and, as such, do not create strong polarization into subcommunities.
This implies that more often than not, there are quite a few specific hot news that all users might express interest in,
and it is natural to expect that these pieces of news are intended to reach a wide audience of consumers.
Given this state of affairs, it is not surprising that on the Yahoo! dataset both {\sc LINUCB-ONE} and {\sc LINUCB-V}
(serving the same news to all users) are already performing quite well, thereby making the clustering-of-users effort somewhat less useful. This also explains the poor performance of {\sc LINUCB-IND}, which is not performing any clustering at all.
Yet, even in this non-trivial case, {\sc COFIBA} can still achieve a significant increased prediction accuracy compared, e.g., to {\sc CLUB}, thereby suggesting that simultaneous clustering at both the user and the item (the news) sides might be an even more effective strategy to earn clicks in news recommendation systems.

Most of the users in the Telefonica data are from a diverse sample of people in Spain, and it is easy to imagine that this dataset spans a large number of communities across its population. Thus we can assume that collaborative effects will be much more evident, and that
{\sc COFIBA} will be able to leverage these effects efficiently. In this dataset, {\sc CLUB} performs well in general,
while {\sc DYNUCB} deteriorates in the initial stage and catches-up later on.
{\sc COFIBA} seems to surpass all other algorithms, especially in the cold-start regime, all other algorithms being in the same ballpark as {\sc CLUB}.
Finally, the Avazu data is furnished from its professional digital advertising solution platform, where the customers click the ad impressions
via the iOS/Android mobile apps or through websites, serving either the publisher or the advertiser which leads to a daily
high volume internet traffic.
In this dataset,
neither {\sc LINUCB-ONE} nor {\sc LINUCB-IND} displayed a competitive cold-start performance. {\sc DYNUCB} is underperforming throughout,
while {\sc LINUCB-V} demonstrates a relatively high CTR. {\sc CLUB} is strong at the beginning, but then its CTR
performance degrades. On the other hand, {\sc COFIBA} seems to work extremely well during the cold-start, and comparatively
best in all later stages.




In Figure \ref{fig:bars} we give a typical distribution of cluster sizes produced by {\sc COFIBA} after at the end of its
run.\footnote
{
Without loss of generality, we take the first Yahoo dataset to provide statistics, for similar shapes of the bar plots can be established for the remaining ones.
}
The emerging pattern is always the same: we have few clusters over the items with very unbalanced sizes and, corresponding to each item cluster, we have few clusters over the users, again with very unbalanced sizes. This recurring pattern is in fact
the motivation behind our theoretical assumptions (Section \ref{s:model}), and a property of data that the {\sc COFIBA} algorithm can provably take advantage of (Section \ref{s:analysis}). These bar plots, combined with the comparatively good performance of {\sc COFIBA}, suggest that our datasets do actually possess clusterability properties at both sides.

To summarize, despite the differences in the three datasets, the experimental evidence we collected on them is quite consistent,
in that in all the three cases {\sc COFIBA} significantly outperforms all other competing methods we tested.
This is especially noticeable during the cold-start period, but the same relative behavior essentially shows up
during the whole time window of our experiments. {\sc COFIBA} is a bit involved to implement, as contrasted to its competitors, and is also somewhat slower to run (unsurprisingly slower than, say, {\sc LINUCB-ONE} and {\sc LINUCB-IND}). On the other hand,
{\sc COFIBA} is far more effective in exploiting the collaborative effects embedded in the data, and still amenable to
be run on large datasets.



\section{Regret Analysis}\label{s:analysis}
The following theorem is the 
theoretical guarantee of {\sc COFIBA},
where we relate the cumulative regret of {\sc COFIBA} to the clustering structure of users $\scU$ w.r.t. items $\scI$.
For simplicity of presentation,
we formulate our result in the one-hot encoding case, where $\bu_i \in \R^d$, $i = 1, \ldots, n$, and $\scI = \{\be_1, \ldots, \be_d\}$. In fact, a more general statement can be proven which holds in the case when $\scI$
is a generic set of feature vectors $\scI = \{\bx_1, \ldots, \bx_{|\scI|}\}$, and the regret bound depends
on the geometric properties of such vectors.\footnote
{
In addition, the function $\CB$ should be modified so as to incorporate these properties.
}

In order to obtain a provable advantage from our clusterability assumptions,
extra conditions are needed on the way $i_t$ and $C_{i_t}$ are generated.
The clusterability assumptions we can naturally take advantage of are those where, for most partitions
$P(\be_h)$, the relative sizes of clusters over users are highly unbalanced.
Translated into more practical terms, cluster unbalancedness amounts to saying that the universe of items $\scI$ tends to
influence users so as to determine a small number of major common behaviors (which need neither be the same
nor involve the same users across items), along with a number of minor ones.
As we saw in our experiments, this seems like a frequent
behavior of users in some practical
scenarios.
\begin{theorem}\label{t:main}
Let the {\sc COFIBA} algorithm of Figure~\ref{alg:cofiba} be run on a set of users $\scU = \{1,\ldots,n\}$
with associated profile vectors $\bu_1, \ldots, \bu_n \in \R^d$, and set of items $\scI = \{\be_1, \ldots, \be_d\}$
such that the $h$-th induced partition $P(\be_h)$ over $\scU$ is made up of $m_h$ clusters of cardinality
$v_{h,1}, v_{h,2}, \ldots, v_{h,m_h}$, respectively.
Moreover, let $g$ be the number of {\em distinct} partitions so obtained.
At each round $t$, let $i_t$ be generated uniformly at random\footnote
{
Any distribution having positive probability on each $i \in \scU$ would suffice here.
}
from $\scU$. Once $i_t$ is selected, the number $c_t$ of items in $C_{i_t}$
is generated arbitrarily as a function of past indices $i_1, \ldots, i_{t-1}$, payoffs $a_1, \ldots, a_{t-1}$,
and sets $C_{i_1}, \ldots, C_{i_{t-1}}$, as well as the current index $i_t$.
Then the sequence of items
in $C_{i_t}$ is generated i.i.d. (conditioned on $i_t$, $c_t$ and all past indices
$i_1, \ldots, i_{t-1}$, payoffs $a_1, \ldots, a_{t-1}$, and sets $C_{i_1}, \ldots, C_{i_{t-1}}$)
according to a given but unknown distribution $\scD$ over $\scI$.
Let payoff $a_t$ lie in the interval $[-1,1]$, and be generated as described in Section \ref{s:model}
so that, conditioned on history, the expectation of $a_t$ is $\bu_{i_t}^\top{\bar \bx_t}$.
Finally, let parameters $\alpha$ and $\alpha_2$ be suitable functions of $\log (1/\delta)$.
%
%
If $c_t \leq c$\, $\forall t$ then, as $T$ grows large,
with probability at least $1-\delta$ the cumulative regret satisfies\footnote
{
The ${\tilde \scO}$-notation hides logarithmic factors in $n$, $m$, $g$, $T$, $d$, $1/\delta$, as well as terms which are independent of $T$.
}
\begin{align*}
\sum_{t=1}^T r_t
= {\tilde \scO}\left(
\left(\E[S]
+ \sqrt{c\,\sqrt{mn}\,\var(S)}+1\right)
\sqrt{\frac{d\,T}{n}}
\right)~,
\label{e:asymptotic_bound}
\end{align*}
where\ $S = S(h) = \sum_{j=1}^{m_h} \sqrt{v_{h,j}}$\,,
$h$ is a random index such that $\be_h \sim \scD$, and $\E[\cdot]$
and $\var(\cdot)$
denote, respectively, the expectation and the variance w.r.t. this random index.
\end{theorem}
To get a feeling of how big (or small) $\E[S]$ and $\var[S]$ can be, let us consider
the case where each partition over users has a single big cluster and a number of small ones.
To make it clear, consider the extreme scenario where each $P(\be_h)$ has one cluster of size $v_{h,1} = n-(m-1)$,
and $m-1$ clusters of size $v_{h,j} = 1$, with $m < \sqrt{n}$. Then it is easy to see that
$\E[S] = \sqrt{n-(m-1)} + m-1$, and $\var(S) =0$,
so that the resulting regret bound essentially becomes ${\tilde \scO}(\sqrt{dT})$,
which is
the standard regret bound one achieves for learning a {\em single} $d$-dimensional user
(aka, the standard noncontextual bandit bound with $d$ actions and no gap assumptions among them).
At the other extreme lies the case when each partition $P(\be_h)$ has $n$-many clusters, so that
$\E[S] = n$, $\var(S) =0$,
and the resulting bound is ${\tilde \scO}(\sqrt{dnT})$. Looser upper bounds
can be achieved in the case when $\var(S) > 0$, where also the interplay with $c$ starts becoming
relevant.
%
%
Finally, observe that the number $g$ of distinct partitions influences the bound only indirectly through
$\var(S)$. Yet, it is worth repeating here that $g$ plays a crucial role in the computational (both time
and space) complexity of the whole procedure.

\begin{proof}[of Theorem \ref{t:main}]
The proof sketch builds on the analysis in \cite{glz14}. Let the true underlying clusters over the users
be $V_{h,1}, V_{h,2}, \ldots, V_{h,m_h}$, with $|V_{h,j}| = v_{h,j}$. In \cite{glz14}, the authors show
that, because each user $i$ has probability $1/n$ to be the one served in round $t$, we have, with high probability, $\bw_{i,t} \rightarrow \bu_i$ for all $i$, as $t$ grows large.
Moreover, because of the gap assumption involving parameter $\gamma$, all edges connecting users belonging
to different clusters at the user side will eventually be deleted (again, with high probability), after each user $i$ is served
at least $O(\frac{1}{\gamma^2})$ times. By the way edges are disconnected at the item side, the above is essentially independent (up to log factors due to union bounds) of which graph at the user side we are referring to.
In turn, this entails that the current user clusters encoded by the connected components of graph $G^U_{t,h}$ will eventually converge to the $m_h$ true user clusters (again, independent of $h$, up to log factors), so that the aggregate weight vectors $\bbw_{N_k,t-1}$ computed by the algorithm for trading off exploration vs. exploitation in round $t$ will essentially converge to $\bu_{i_t}$ at a rate of the form\footnote
{
Because $\scI = \{\be_1, \ldots, \be_d\}$, the minimal eigenvalue $\lambda$ of the process correlation matrix $\E[X\,X^\top]$ in \cite{glz14} is here $1/d$. Moreover, compared to \cite{glz14}, we do not strive to capture the geometry of the user vectors $\bu_i$ in the regret bound, hence we do not have the extra $\sqrt{m}$ factor occurring in their bound.
}
\begin{equation}\label{e:rate}
\E\left[\frac{1}{\sqrt{1+T_{h_t,j_t,t-1}/d}}\right]~,
\end{equation}
where $h_t$ is the index of the true cluster over {\em items} that ${\bar \bx_t}$ belongs to, $j_t$ is the index of the true cluster over {\em users} that $i_t$ belongs to (according to the partition of $\scU$ determined by $h_t$), $T_{h_t,j_t,t-1}$ is the number of rounds so far where we happened to ``hit" cluster $V_{h_t,j_t}$, i.e.,
\[
T_{h_t,j_t,t-1} = |\{s \leq t-1\,:\,i_s \in V_{h_t,j_t} \}|~,
\]
and the expectation is w.r.t. both the (uniform) distribution of $i_t$, and distribution $\scD$ generating the items
in $C_{i_t}$, conditioned on all past events. Since, by the Azuma-Hoeffding inequality, $T_{h_t,j_t,t-1}$ concentrates as
\vspace{-0.02in}
\[
T_{h_t,j_t,t-1} \approx \frac{t-1}{n}\,v_{h_t,j_t}~,
\]
we have
%
\vspace{-0.1in}
\[
(\ref{e:rate}) \approx \E_{\scD}\left[ \sum_{j=1}^{m_{h_t}} \frac{v_{h_t,j}}{n}\,\frac{1}{\sqrt{1+\frac{t-1}{d\,n}\,v_{h_t,j}}} \right]~.
\]
%
It is the latter expression that rules the cumulative regret of {\sc COFIBA} in that, up to log factors:
\begin{equation}\label{e:rate3}
\sum_{t=1}^T r_t
\approx \sum_{t=1}^T \E_{\scD}\left[ \sum_{j=1}^{m_{h_t}} \frac{v_{h_t,j}}{n}\,\frac{1}{\sqrt{1+\frac{t-1}{d\,n}\,v_{h_t,j}}} \right]~.
\end{equation}
Eq. (\ref{e:rate3}) is essentially (up to log factors and omitted additive terms) the regret bound one would obtain by knowning beforehand the latent clustering structure over $\scU$.

Because $h_t \in C_{i_t}$ is itself a function of the items in $C_{i_t}$, we can eliminate the dependence on $h_t$ by the following simple stratification argument. First of all, notice that
\[
\sum_{j=1}^{m_{h_t}} \frac{v_{h_t,j}}{n}\,\frac{1}{\sqrt{1+\frac{t-1}{d\,n}\,v_{h_t,j}}}
\approx
\sqrt{\frac{d}{nt}}\,\sum_{j=1}^{m_{h_t}} \sqrt{v_{h_t,j}}~.
\]
Then, we set for brevity
$S(h) = \sum_{j=1}^{m_{h}} \sqrt{v_{h,j}}$,
and let $h_{t,k}$ be the index of the true cluster over items that $\bx_{t,k}$ belongs to (recall that $h_{t,k}$ is a random variable since so is $\bx_{t,k}$). Since $S(h_{t,k}) \leq \sqrt{mn}$,
a standard argument shows that
\begin{align*}
\E_{\scD}\left[ S(h_t) \right] &\leq \E_{\scD}\left[ \max_{k=1,\ldots,c_t} S(h_{t,k}) \right] \\
&\leq
\E_{\scD}[S(h_{t,1})] + \sqrt{c\,\sqrt{mn}\,\var_{\scD}(S(h_{t,1}))}+1~,
\end{align*}
so that, after some overapproximations,
we conclude that $\sum_{t=1}^T r_t$ is upper bounded with high probability by
\begin{align*}
{\tilde \scO}\left(\left(\E_{\scD}[S(h)]
+ \sqrt{c\,\sqrt{mn}\,\var_{\scD}(S(h))}+1\right)
\sqrt{\frac{d\,T}{n}}\right)~,
\end{align*}
the expectation and the variance being over the random index $h$ such that $\be_h \sim \scD$.
%
\end{proof}

\section{Conclusions}
We have initiated an investigation of collaborative filtering bandit algorithms operating in relevant scenarios
where multiple users can be grouped by behavior similarity in different ways w.r.t.
items and, in turn, the universe of items can possibly be grouped
by the similarity of clusterings they induce over users. We carried out an extensive experimental
comparison with very encouraging results, and have also given a regret analysis which operates in a simplified scenario.
Our algorithm can in principle be modified so as to be combined with any standard clustering (or co-clustering) technique.
However, one advantage of encoding clusters as connected components of graphs (at least at the user side) is that we are quite effective in tackling the so-called {\em cold start} problem, for the newly served users are more likely
to be connected to the old ones, which makes {\sc COFIBA} in a position to automatically propagate information from the old users
to the new ones through the aggregate vectors $\bbw_{N_k,t}$.
In fact, so far we have not seen any other way of adaptively clustering users and items which is
computationally affordable on sizeable datasets and, at the same time, amenable to a regret analysis
that takes advantage of the clustering assumption.

All our experiments have been conducted in the setup of one-hot encoding, since the datasets
at our disposal did not come with reliable/useful annotations on data. Yet, the
algorithm we presented can clearly work when the items are accompanied by (numerical) features.
One direction of our future research is to compensate for the lack of features
in the data by first {\em inferring} features during an initial training phase through standard matrix
factorization techniques, and subsequently applying our algorithm to a universe of
items $\scI$ described through such inferred features.
Another line of experimental research would be to combine different bandit algorithms (possibly at
different stages of the learning process) so as to roughly get the best of all of them in all stages.
This would be somewhat similar to the meta-bandit construction described in \cite{tjll14}. Another one would be to combine with matrix factorization techniques as in, e.g., \cite{kbktc15}.

\section{Acknowledgments}
We would like to thank the anonymous reviewers for their helpful and constructive comments. The first author thanks the support from MIUR and QCRI-HBKU. Also, the first and the third author acknowledge the support from Amazon AWS Award in Machine Learning Research Grant. The work leading to these results has received funding from the European Union's Seventh Framework Programme (FP7/
2007-2013) under CrowdRec Grant Agreement n$^\circ$ 610594.
\bibliographystyle{abbrv}
\begin{small}
  \vspace{0.3cm}

\end{small}  
%
%

\end{document}